\documentclass[10pt,conference,letterpaper]{IEEEtran}

\IEEEoverridecommandlockouts

\usepackage{times,amsmath,epsfig,amssymb,graphicx,amsfonts,amsthm,dsfont}
\usepackage{subfigure}
\usepackage{setspace}
\usepackage{graphicx}
\usepackage{empheq}
\usepackage{color}
\usepackage{psfrag}
\usepackage{fge}
\usepackage{amsmath}
\usepackage{amsthm}
\usepackage{amsfonts}   
\usepackage{amssymb}   
\usepackage{chemarrow} 
\usepackage{mathrsfs}
\usepackage{setspace}
\usepackage{algorithm}
\usepackage{algcompatible}
\usepackage{lipsum}
\usepackage{mathrsfs}%
\usepackage{mathtools}
\mathtoolsset{showonlyrefs}
\usepackage{bbm}
\floatname{algorithm}{Routine}

\newtheorem{mydef}{Definition}

\newtheorem{mytheorem}{Theorem}
\newtheorem{mylemma}{Lemma}

\newtheorem{myproposition}{Proposition}

\newtheorem{myexample}{Example}

\newcounter{ale}

\newenvironment{liste}{\begin{itemize}}{\end{itemize}}
\newcommand{\aliste}{\begin{liste} \setcounter{ale}{1}}
\newcommand{\zliste}{\end{liste}}

\begin{document}

\title{{Switching to Learn}} 

\author{Shahin Shahrampour, Mohammad Amin Rahimian, Ali Jadbabaie$^*$
\thanks{$^*$ The authors are with the Department of Electrical and Systems Engineering
and General Robotics, Automation, Sensing and Perception (GRASP)
Laboratory, University of Pennsylvania, Philadelphia, PA 19104-6228 USA (emails: {\tt\small \{shahin,mohar,jadbabai\}@seas.upenn.edu}). This work was supported by ARO MURI W911NF-12-1-0509.}

}


\maketitle

\begin{abstract}
A network of agents attempt to learn some unknown state of the world drawn by nature from a finite set. Agents observe private signals conditioned on the true state, and form beliefs about the unknown state accordingly. Each agent may face an identification problem in the sense that she cannot distinguish the truth in isolation. However, by communicating with each other, agents are able to benefit from side observations to learn the truth collectively. Unlike many distributed algorithms which rely on all-time communication protocols, we propose an efficient method by switching between Bayesian and non-Bayesian regimes. In this model, agents exchange information only when their private signals are not informative enough; thence, by switching between the two regimes, agents efficiently learn the truth using only a few rounds of communications. The proposed algorithm preserves learnability while incurring a lower communication cost. We also verify our theoretical findings by simulation examples.
\end{abstract}

\section{Introduction}\label{sec:intro}


Distributed estimation, detection, and learning theory in networks have attracted much attention over the past decades \cite{BorVar82,Tsit88,tsitsiklis1993decentralized,jadbabaie2003coordination}, 
with applications that range from sensor and robotic networks \cite{Olfati05,tsitsiklis1984problems,kar2012distributed,atanasov2014distributed,7040469} to social and economic networks \cite{jadbabaie2012non,shahrampour2013online,MosselSlyTamuz14}.  In these scenarios, agents in a network need to learn the value of a parameter that they may not be able to infer on their own, but the global spread of information in the network provides them with adequate data to learn the truth collectively. As a result, agents iteratively exchange information with their neighbors. For instance, in distributed sensor and robotic networks, agents use local diffusion to augment their imperfect observations with information from their neighbors and achieve consensus and coordination~\cite{bullo2009distributed,EgerMesBook}. Similarly, agents exchange beliefs in social networks to benefit from each other's observations and private information and learn the unknown state of the world \cite{Chamley2004,Jackson2008}. 

Existing literature on distributed learning focuses mostly on environments where individuals communicate at every round. Of particular relevance to our discussion are a host of algorithms that follow the non-Bayesian learning scheme in Jadbabaie et. al. \cite{jadbabaie2012non}. In their seminal work, the authors propose an observational social learning model using purely local diffusions. At any round, each agent performs a Bayesian update based on her privately observed signal and uses a linear convex combination to incorporate her Bayesian posterior with beliefs of her neighbors and obtain a refined opinion. Inspired by \cite{jadbabaie2012non}, many algorithms are developed that either rely on all-time communication protocols \cite{rad2010distributed,shahrampour2014distributed,lalitha2014social,rahimian2014non} or follow structured switching rules \cite{shahin_CDC_2013,nedic2014nonasymptotic}. For instance, in \cite{shahin_CDC_2013} Shahrampour et. al. propose a scheme based on a gossip algorithm, and in \cite{nedic2014nonasymptotic} Nedi{\'c} et. al. present a method effective for switching topologies which respect a set of assumptions. 

The chief aim of this note is to  consider a scenario where communication at any given time $t$ occurs only if an agent's belief does not change drastically due to her private observation at that time $t$; i.e. the agent's private signal is not \emph{informative} enough. Accordingly, an agent uses the Bayes' rule to update her belief with every \emph{strong} private signal that she observes; otherwise, she uses a non-Bayesian averaging rule to refine her opinion, by incorporating her neighbors' observations and own private signals and in a non-Bayesian manner.


Our contributions are as follows. we propose the \emph{total variation distance} between the current belief of each agent and the Bayesian update after observing a given signal, as the criterion for characterization of informativeness. In particular, a private signal is deemed informative, if the distance between the agent's current belief and her Bayesian posterior given her private signal exceeds a preset threshold. Given the proposed criterion for informativeness, we implement a switching mechanism with agents shifting from Bayesian to non-Bayesian regime and vice versa. In the Bayesian regime, every agent uses the Bayes' rule to update her belief based on her privately observed signal. In the non-Bayesian regime, the agents  use an averaging rule to combine the observations communicated by their neighbors with their private signals. The challenge of analysis is due to the fact that the network topology becomes a function of signals, and does not evolve independently across time. Under some mild assumptions, we are able to show that by switching between the two regimes based on the informativeness of signals, agents can efficiently learn the truth. We further provide an asymptotic rate of convergence, and discuss the performance of the algorithm in numerical experiments.

The remainder of this paper is organized as follows. The problem formulation and modeling details are set forth in Section~\ref{sec:pre}. The main results are presented in Section~\ref{sec:main}, where we begin by describing the characterization of informativeness and the proposed switching rules in Subsections~\ref{sec:informative} and \ref{sec:switching}, respectively; followed by the convergence analysis in Subsection~\ref{sec:rate}. Simulation examples and discussions in Section \ref{sec:sim} illustrate the results. Concluding remarks and the future directions are provided in Section~\ref{sec:conc}. All proofs are included in the appendix.

\section{Problem Formulation}\label{sec:pre}

\paragraph*{Notation}  Throughout, $\mathbb{R}$ is the set of real numbers, $\mathbb{N}$ denotes the set of natural numbers, and $\mathbb{W} = \mathbb{N}\cup \{0\}$. For any fixed integer $n \in \mathbb{N}$ the set of integers $\{1,2,\ldots,n\}$ is denoted by $[n]$, while any other set is represented by a calligraphic capital letter. The cardinality of a set $\mathcal{X}$, which is the number of its elements, is denoted by $|\mathcal{X}|$; and $\mathcal{P}(\mathcal{X})$ is the power-set of $\mathcal{X}$, which is the set of all its subsets. Boldface letters denote random variables, and vectors are in column form. $I_n$ denotes the $n\times n$ identity matrix, $\mathds{1}$ represents the vector of all ones, and $^T$ denotes the matrix transpose. 

\subsection{The Model}

Consider a set of $n$ agents that are labeled by $[n]$ and interact according to a weighted and directed graph $\mathcal{G} = ([n],\mathcal{E},P)$, where $\mathcal{E} \subseteq [n] \times [n]$ is the set of edges and $P \in \mathbb{R}^{n \times n}$ is a symmetric doubly stochastic matrix. The $ij$-th entry of $P$, denoted by $p_{ij} = [P]_{ij}$, assigns a positive weight to edge $(i,j)$ if $(i,j) \in \mathcal{E}$, and sets $p_{ij} = 0$ if $(i,j) \not\in \mathcal{E}$. We further have $[P]_{ii}>0$ for every $i \in [n]$, i.e., all agents have positive \emph{self-reliant}. $\mathcal{N}(i) = \{j \in [n]; (j,i) \in \mathcal{E} , j\neq i\}$ is called the neighborhood of agent $i$. 

The goal of each agent is to decide between one of the $m$ possible states from the state space $\Theta$. $\Delta\Theta$ is the space of all probability measures on the set $\Theta$. A random variable $\boldsymbol{\theta}$ is chosen randomly from $\Theta$ by nature and according to the probability measure $\nu(\mathord{\cdot}) \in \Delta\Theta$, which satisfies $\nu(\hat{\theta}) > 0$ for all $\hat{\theta} \in \Theta$, and is referred to as the common prior. Associated with each agent $i$, $\mathcal{S}_i$ is a finite set called the signal space of $i$, and given $\boldsymbol{\theta}$, $\ell_i(\mathord{\cdot}|\boldsymbol{\theta})$ is a probability measure on $\mathcal{S}_i$, which is referred to as the \emph{signal structure} or \emph{likelihood function} of agent $i$. Furthermore, $(\Omega,\mathscr{F},\mathbb{P})$ is a probability triple, where $\Omega = \otimes_{t=0}^\infty {\left(\prod_{i\in[n]}\mathcal{S}_i\right)} \times \Theta$ is an infinite product space with a general element $\omega = ((s_{1,0},\ldots,s_{n,0}),(s_{1,1},\ldots,s_{n,1}),\ldots;\theta)$ and the associated sigma field $\mathscr{F} = \mathcal{P}(\Omega)$.  $\mathbb{P}(\mathord{\cdot})$ is the probability measure on $\Omega$  which assigns probabilities consistently with the common prior $\nu(\mathord{\cdot})$ and the likelihood functions $\ell_i(\mathord{\cdot}|\boldsymbol{\theta}), i \in [n]$, such that conditioned on $\boldsymbol{\theta}$ the random variables $\{\mathbf{s}_{i,t},i\in[n],t\in\mathbb{W}\}$ are independent. Note that the observed signals are independent and identically distributed over time, and independent across the agents at each epoch of time. $\mathbb{E}[\mathord{\cdot}]$ is the expectation operator, which represents integration with respect to $\mathrm{d}\mathbb{P}(\omega)$. Let $\theta$ be the unknown state drawn initially by nature. Since signals are generated based on $\theta$, we have that
\begin{align}
\mathbb{E}\left[\log \frac{\ell_i(\cdot\vert \hat{\theta})}{\ell_i(\cdot\vert \theta)} \right] =  - D_{KL}\left(\ell_i( \mathord{\cdot} |\theta) \| \ell_i( \mathord{\cdot} |\hat{\theta}) \right) \leqslant 0,
\end{align} where the inequality follows from the fact that $D_{KL}\left(\mathord{\cdot}\| \mathord{\cdot} \right)$, the Kullback-Leibler divergence, is always nonnegative \cite{CoverThomas1991}. The inequality is strict if and only if $\ell_i( \mathord{\cdot} |\hat{\theta}) \not\equiv \ell_i( \mathord{\cdot} |\theta)$, i.e. $\exists s \in \mathcal{S}_i$ such that $\ell_i( s |\hat{\theta}) \neq \ell_i( s |\theta)$. Note that whenever $\ell_i( \mathord{\cdot} |\hat{\theta}) \equiv \ell_i( \mathord{\cdot} |\theta)$ or equivalently $D_{KL}\left(\ell_i( \mathord{\cdot} |\hat{\theta})\| \ell_i( \mathord{\cdot} |\theta)) \right) = 0$, then the two states $\hat{\theta}$ and $\theta$ are statically indistinguishable to agent $i$. In other words, there is no way for agent $i$ to differentiate $\hat{\theta}$ from $\theta$ based only on her private signals. This follows from the fact that both $\theta$ and $\hat{\theta}$ induce the same probability distribution on her sequence of observed i.i.d. signals. We, therefore, have the following characterization.

\begin{mydef}[Observationally Equivalent States]\label{def:observationallyEquivalnet} For any $\hat{\theta} \in \Theta$ the set of states $\tilde{\theta}\in\Theta$ that are observationally equivalent to $\hat{\theta}$ for agent $i$ are given by $\mathcal{O}_{i}(\hat{\theta}) = \left\{\tilde{\theta}\in\Theta  : \ell_i( \mathord{\cdot} |\hat{\theta}) \equiv \ell_i( \mathord{\cdot} |\tilde{\theta})  \right\}$.
\end{mydef}
To distinguish the true state of the world $\theta$ from any false state $\hat{\theta}\neq\theta$, there must exist an agent that is able to detect $\hat\theta$ as a false state, in which case it holds that \cite{shahrampour2014distributed}
\begin{align}
\mathcal{I}(\hat{\theta},\theta)  := -\frac{1}{n}\sum_{i=1}^{n}  D_{KL}\left(\ell_i( \mathord{\cdot} |\theta) \| \ell_i( \mathord{\cdot} |\hat{\theta}) \right) < 0.
\label{GAMMAdefinition}
\end{align} 
We, therefore, have the following characterization.

\begin{mydef}[Globally Identifiability]\label{def:learnability} The true state $\theta$ is globally identifiable, if  $\mathcal{I}(\hat{\theta},\theta)< 0$ for all $\hat{\theta}$ $\in$ $\Theta\fgebackslash \{\theta\}$.
\end{mydef}
We adhere to the following assumptions throughout the paper.
\begin{description}
\item[{\bf A1.}] All log-marginals are uniformly bounded such that $|\log \ell_i(s_i |\hat{\theta})| \leq B$ for all $i \in [n]$, $s_i \in \mathcal{S}_i$, and any $\hat{\theta}\in \Theta$.
\item[{\bf A2.}] The true state is \emph{globally identifiable}, i.e., we have $\mathcal{I}(\hat{\theta},\theta)<0$ for any $\hat{\theta}$ $\in$ $\Theta\fgebackslash \{\theta\}$.
\item[{\bf A3.}] The graph $\mathcal{G}$ is \emph{strongly connected}, i.e., there exists a directed path from any node $i\in [n]$ to any node $j\in [n]$. 
\end{description}
Assumption {\bf A1} implies that every signal has a bounded information content. For instance, it holds when the signal space is discrete. Assumption {\bf A2} guarantees that accumulation of likelihoods provides sufficient information to make the true state uniquely identifiable from the aggregate observations of all agents across the network. Finally, the strong connectivity (assumption {\bf A3}) guarantees the information flow in the network. 
We end this section by the following definition \cite{moreau2005stability}.
\begin{mydef}[Connectivity]\label{def:connectivity}
Consider a sequence of directed graphs $\mathcal{G}_t = ([n],\mathcal{E}_t,A_t)$ for $t\in \mathbb{N}$, where $A_t$ is a stochastic matrix. A node $i\in [n]$ is connected to a node $j\neq i$ across an interval $\mathcal{T}\subseteq \mathbb{N}$ if there exists a directed path from $i$ to $j$ for the directed graph $([n],\cup_{t\in \mathcal{T}} \mathcal{E}_t)$.
\end{mydef}

\subsection{Belief Updates}

For each time instant $t$, let ${\boldsymbol\mu}_{i,t}(\mathord{\cdot})$ be the probability mass function on $\Theta$, representing the \emph{opinion} or \emph{belief} at time $t$ of agent $i$ about the unknown state of the world. The goal is to investigate the problem of asymptotic learning, that is each agent learning the true realized value $\theta$. The convergence can be in the probability or almost sure sense. In this paper, we are interested in asymptotic and almost sure characterization of learning, formalized as follows.
\begin{mydef}[Learning] An agent $i \in [n]$ learns the true state $\theta$ asymptotically, if ${\boldsymbol\mu}_{i,t}(\theta) \longrightarrow 1$, $\mathbb{P}$-almost surely.
\end{mydef}

At $t= 0$, the value $\boldsymbol{\theta} = \theta$ is realized, and each agent $i\in [n]$ forms an initial Bayesian opinion ${\boldsymbol\mu}_{i,0}(\mathord{\cdot})$ about the value of $\boldsymbol{\theta}$. Given the signal $\mathbf{s}_{i,0}$, and using Bayesian update for each agent $i\in[n]$, her initial belief in terms of the observed signal $\mathbf{s}_{i,0}$ is given by,
\begin{equation}
{\boldsymbol\mu}_{i,0}(\hat{\theta}) = \frac{ \nu(\hat{\theta})\ell_i(\mathbf{s}_{i,0}|\hat{\theta})} {\sum_{\tilde{\theta} \in \Theta}\nu(\tilde{\theta})\ell_i(\mathbf{s}_{i,0}|\tilde{\theta})},\ \ \forall \hat{\theta} \in \Theta.
\label{eq:bayes1}
\end{equation} 
At any $t \in \mathbb{N}$, agent $i$ uses the following update rule to calculate $\boldsymbol{\phi}_{i,t}(\hat{\theta})$, 
\begin{align}
\boldsymbol{\phi}_{i,t}(\hat{\theta}) =  \label{eq:nonBayesianUpdate} \sum_{j=1}^{n}[\mathbf{Q}_t]_{ij}\boldsymbol{\phi}_{j,t-1}(\hat{\theta}) + \log \ell_{i}(\mathbf{s}_{i,t}\vert\hat{\theta}),
\end{align}
for any $\hat{\theta}\in \Theta$, where $\mathbf{Q}_t$ is a real $n \times n$ matrix (possibly random and time varying) and $\boldsymbol\phi_{i,0}(\hat{\theta})=0$ by convention. Then she updates her belief $\boldsymbol\mu_{i,t}(\hat{\theta})$ as 
\begin{align}\label{nonbayes}
\boldsymbol\mu_{i,t}(\hat{\theta}) = \frac{\boldsymbol\mu_{i,0}(\hat{\theta})e^{\boldsymbol\phi_{i,t}(\hat{\theta})}}{\sum_{\tilde{\theta}\in\Theta}\boldsymbol\mu_{i,0}(\tilde{\theta})e^{\boldsymbol\phi_{i,t}(\tilde{\theta})}}.
\end{align}
for any $\hat{\theta}\in \Theta$. In section \ref{sec:switching}, we shall describe in detail the switching strategy under which $\mathbf{Q}_t$ evolves. If there is no communication among agents, we have $\mathbf{Q}_t = I_n$. Hence, each agent $i$ observes the realized value of $\mathbf{s}_{i,t}$, calculates the likelihood $\ell_i(\mathbf{s}_{i,t}|\hat{\theta})$ for any $\hat{\theta} \in \Theta$, and forms an opinion using the Bayes' rule
\begin{align}
{\boldsymbol\mu}^{B}_{i,t}(\hat{\theta}) = \frac{ {\boldsymbol\mu}_{i,t-1}(\hat{\theta})\ell_i(\mathbf{s}_{i,t}|\hat{\theta})} {\sum_{\tilde{\theta} \in \Theta}{\boldsymbol\mu}_{i,t-1}(\tilde{\theta})\ell_i(\mathbf{s}_{i,t}|\tilde{\theta})},
\label{eq:BayesianUpdate}
\end{align}
where ${\boldsymbol\mu}_{i,t-1}(\mathord{\cdot})$ is calculated using \eqref{nonbayes}. Alternatively, at any time $t$ that the Bayes' update based on the private signal $\mathbf{s}_{i,t}$ does not provide enough information (on which we elaborate in section \ref{sec:switching}), agent $i$ switches to a non-Bayesian update, incorporating her neighboring beliefs but only for that particular unit of time $t$. Collecting log-likelihoods from her neighborhood, agent $i\in [n]$ averages the local data by performing \eqref{eq:nonBayesianUpdate} with $[\mathbf{Q}_t]_{ij} = [P]_{ij}$ and uses the resultant $\boldsymbol{\phi}_{i,t}(\hat{\theta})$ in \eqref{nonbayes} to obtain a refined but non-Bayesian opinion $\boldsymbol{\mu}_{i,t}(\mathord{\cdot})$. One can view the learning rules \eqref{eq:nonBayesianUpdate} and \eqref{nonbayes} for each agent $i$, as repeated Bayesian updates in an infinite sequence of contiguous, nonempty and bounded time-intervals. At the outset of each interval, the agent's prior is derived based on averaging the local information from her neighbors, while during the interval there is no communication and the agent performs successive Bayesian updates based on her private signals. Writing the matrix form of \eqref{eq:nonBayesianUpdate}, it can be  verified (see Lemma 3 in \cite{shahin_CDC_2013}) that 
\begin{align}\label{therightphi}
\boldsymbol\phi_{i,t}(\hat{\theta}) = \sum_{\tau=0}^{t}\sum_{j=1}^n\left[\prod_{\rho=0}^{t-1-\tau}\mathbf{Q}_{t-\rho}\right]_{ij}\log \ell_j(\mathbf{s}_{j,\tau} |\hat{\theta}).
\end{align}

Finally, note that choosing $\mathbf{Q}_t$ at each time $t$ based on a gossip protocol reduces the setting to \cite{shahin_CDC_2013}, while $\mathbf{Q}_t=P$ recovers the model considered in \cite{shahrampour2014distributed}. In both cases convergence of beliefs occurs by incurring the cost of communicating at every round.

\section{Main Results}\label{sec:main}

In this section, we propose the switching rule based on which the (possibly random and time varying) matrix $\mathbf{Q}_t$ in \eqref{eq:nonBayesianUpdate} is chosen. The rule characterizes the dichotomy between the non-communicative Bayesian and communicative non-Bayesian regime. We shall prove that all agents learn the truth efficiently under this protocol. The switching rule, as we describe next, occurs based on the quality of information that private signals offer. 

\subsection{Characterizing the Class of Informative Signals}\label{sec:informative}

An informative signal is one that substantially influences an agent's opinion. Here we propose the total variation distance between ${\boldsymbol\mu}^{B}_{i,t}(\cdot)$ and ${\boldsymbol\mu}_{i,t-1}(\cdot)$ as the measure of informativeness for a private signal $\mathbf{s}_{i,t}$. In particular, the private signal $\mathbf{s}_{i,t}$ is \emph{informative} for agent $i$ at time $t$ if $\|{\boldsymbol\mu}^{B}_{i,t}(\cdot)-{\boldsymbol\mu}_{i,t-1}(\cdot) \|_{TV} \geqslant \tau$, where $0 < \tau \leq 1$ is a given threshold. 

\begin{myexample}Informative Signals in a Binary World\end{myexample}
Consider the case where $\Theta = \{1,2\}$, and the true state is $\theta = 1$. Define $\boldsymbol\epsilon_{i,t} = \boldsymbol\mu_{i,t}(2)$ as the mass assigned to the false state by agent $i$ at time $t$. For the case of binary state space considered here, the evolution of each agent's beliefs is uniquely characterized by that of $\boldsymbol\epsilon_{i,t}$ and the focus of interest is therefore to have $\boldsymbol\epsilon_{i,t}$ converge to zero almost surely.

Let $r(\mathbf{s}_{i,t}):= \ell_i(\mathbf{s}_{i,t}|1)/\ell_i(\mathbf{s}_{i,t}|2)$ be the {\it likelihood ratio} under signal $\mathbf{s}_{i,t}$. To investigate the conditions for informativeness on the private signals, we start by simplifying the  expression for $\|{\boldsymbol\mu}^{B}_{i,t}(\cdot)-{\boldsymbol\mu}_{i,t-1}(\cdot) \|_{TV}$ as follows:
\begin{align}
&\left\|{\boldsymbol\mu}^{B}_{i,t}(\cdot)-{\boldsymbol\mu}_{i,t-1}(\cdot) \right\|_{TV} =\frac{1}{2}\left\|{\boldsymbol\mu}^{B}_{i,t}(\cdot)-{\boldsymbol\mu}_{i,t-1}(\cdot)\right\|_{1}  \\
&~~~~~~~~~~~~~~~~=\frac{\boldsymbol\epsilon_{i,t-1}(1-\boldsymbol\epsilon_{i,t-1})\big|\ell_i(\mathbf{s}_{i,t} | 1)-\ell_i(\mathbf{s}_{i,t}| 2)\big|}{(1-\boldsymbol\epsilon_{i,t-1})\ell_i(\mathbf{s}_{i,t}|1)+\boldsymbol\epsilon_{i,t-1}\ell_i(\mathbf{s}_{i,t}|2)} \\
&~~~~~~~~~~~~~~~~=\frac{\boldsymbol\epsilon_{i,t-1}(1-\boldsymbol\epsilon_{i,t-1})\big|r(\mathbf{s}_{i,t})-1\big|}{(1-\boldsymbol\epsilon_{i,t-1})r(\mathbf{s}_{i,t})+\boldsymbol\epsilon_{i,t-1}}.
\end{align}
To investigate the informativeness condition $\|{\boldsymbol\mu}^{B}_{i,t}(\cdot)-{\boldsymbol\mu}_{i,t-1}(\cdot) \|_{TV} \geqslant \tau$, we distinguish two cases $r(\mathbf{s}_{i,t})\geqslant 1$ and $r(\mathbf{s}_{i,t})< 1$. For $r(\mathbf{s}_{i,t})\geqslant 1$, we get
\begin{align}
&\|{\boldsymbol\mu}^{B}_{i,t}(\cdot)-{\boldsymbol\mu}_{i,t-1}(\cdot) \|_{TV} \geqslant \tau  \Longleftrightarrow \label{eq:informative1} \\ &\frac{\boldsymbol\epsilon_{i,t-1} (1-\boldsymbol\epsilon_{i,t-1})\big(r(\mathbf{s}_{i,t})-1\big)}{(1-\boldsymbol\epsilon_{i,t-1})r(\mathbf{s}_{i,t})+\boldsymbol\epsilon_{i,t-1}}\geqslant \tau \Longleftrightarrow \\ 
& r(\mathbf{s}_{i,t}) \geq \frac{\tau \boldsymbol\epsilon_{i,t-1}+\boldsymbol\epsilon_{i,t-1}(1-\boldsymbol\epsilon_{i,t-1})}{\boldsymbol\epsilon_{i,t-1}(1-\boldsymbol\epsilon_{i,t-1})-\tau(1-\boldsymbol\epsilon_{i,t-1})},
\end{align} provided that $\boldsymbol\epsilon_{i,t-1} >\tau$; otherwise when $\boldsymbol\epsilon_{i,t-1} \leq\tau$ no signal with a likelihood ratio $r(\mathbf{s}_{i,t})\geqslant 1$ will be regarded as informative. In other words, for an agent whose belief is already sufficiently close to the truth such likely signals are not surprising. 

On the other hand, for $r(\mathbf{s}_{i,t})< 1$ we have,

\begin{align}
&\|{\boldsymbol\mu}^{B}_{i,t}(\cdot)-\boldsymbol\mu_{i,t-1}(\cdot)\|_{TV} \geqslant \tau \Longleftrightarrow \label{eq:informative2} \\ 
&\frac{\boldsymbol\epsilon_{i,t-1}(1-\boldsymbol\epsilon_{i,t-1})\big(1-r(\mathbf{s}_{i,t})\big)}{(1-\boldsymbol\epsilon_{i,t-1})r(\mathbf{s}_{i,t})+\boldsymbol\epsilon_{i,t-1}}\geqslant \tau  \Longleftrightarrow \\ 
&r(\mathbf{s}_{i,t}) \leqslant \frac{\boldsymbol\epsilon_{i,t-1}(1-\boldsymbol\epsilon_{i,t-1})-\tau \boldsymbol\epsilon_{i,t-1}}{\boldsymbol\epsilon_{i,t-1}(1-\boldsymbol\epsilon_{i,t-1})+\tau(1-\boldsymbol\epsilon_{i,t-1})},
\end{align}
when $\boldsymbol\epsilon_{i,t-1} <1 - \tau$; however, an agent whose belief satisfies  $\boldsymbol\epsilon_{i,t-1} \geq 1 - \tau$ has become almost certain on a falsity; whence she finds no signal with $r(\mathbf{s}_{i,t})< 1$ surprising or informative. The preceding conditions characterize the criterion under which agent $i$ regards an observation $\mathbf{s}_{i,t}$ as informative, for a binary state space $\Theta =\{1,2\}$.   $\hfill \blacksquare $

\subsection{The Switching Rule}\label{sec:switching}

Based on the characterization of the informative signals in the previous section, we now introduce a switching strategy. At each epoch $t$, any agent $i\in [n]$ that receives an uninformative private signal $\mathbf{s}_{i,t}$ exchanges her log-marginal with all her neighbors $j\in \mathcal{N}(i)$. On the other hand, if $\mathbf{s}_{i,t}$ is informative for agent $i$, but a set $\mathcal{M}_t\subseteq \mathcal{N}(i)$ of neighboring agents request for information exchange (i.e. signal $\mathbf{s}_{j,t}$ is not informative for neighbor $j$, for all $j\in \mathcal{M}_t$), then agent $i$ exchanges her log-marginal only with those particular neighbors $j \in \mathcal{M}_t$, who are requesting it (and have received uninformative signals). Therefore, the communication is bidirectional, and we have $[\mathbf{Q}_t]_{ij} = [\mathbf{Q}_t]_{ji} = [P]_{ji}$, whenever any or both of the agents $i$ and $j$ have received uninformative signals. Moreover, $[\mathbf{Q}_t]_{ii} = 1- \sum_{j\in \mathcal{M}_t}[\mathbf{Q}_t]_{ij}$, $\forall i$. In particular, whenever all private signals are informative, agents stick to their Bayesian updates \eqref{eq:BayesianUpdate}, and $\mathbf{Q}_t = I_n$. Accordingly, at each time $t$ the weighting matrix $\mathbf{Q}_t$ which appeared in \eqref{eq:nonBayesianUpdate} is a symmetric and doubly stochastic matrix, and we have the following switching rule for any $t\in\mathbb{W}$:

\label{SWITCHRULE}
{\bf Switching Rule :} {\it Given $\tau>0$, for any $i\in [n]$ that satisfies $\|{\boldsymbol\mu}^{B}_{i,t}(\cdot)-{\boldsymbol\mu}_{i,t-1}(\cdot) \|_{TV} < \tau$, the $i$-th column and row of $\mathbf{Q}_t$ take the values of the $i$-th column and row of the symmetric matrix $P$. Then, the diagonal elements of $\mathbf{Q}_t$ are filled such that the matrix is doubly stochastic.}



Before shifting focus to the convergence analysis under the proposed rule, we note that with $\tau = 1$ all signals will be considered uninformative to all agents at every epoch of time; hence, at every time step agents choose to communicate, $\mathbf{Q}_t = P$ $\forall t$, and they learn the truth exponentially fast \cite{shahrampour2014distributed}. However, the learning occurs under an all-time communication protocol, which is inefficient when communication is costly. We shall demonstrate that the same learning quality can be achieved through the proposed switching rule, while incurring only a few rounds of communications.

\subsection{Consensus on the True State}\label{sec:rate}

We now state the technical results of the paper, and provide the proofs in Appendix. The following lemma concerns the behavior of agents in the Bayesian regime. In particular, it guarantees that with probability one, if the switching condition is satisfied at some time $\mathbf{t}_1$, there exists a $\mathbf{t}_2>\mathbf{t}_1$ at which the switching condition is satisfied again. Furthermore, the length of interval $\mathbf{t}_2-\mathbf{t}_1$ is finite almost surely.
\begin{mylemma}[Bayesian Learning]\label{Main Lemma} Let the log-marginals be bounded (assumption {\bf A1}). Assume that agent $i \in [n]$ is allowed to follow the Bayesian update \eqref{eq:BayesianUpdate} after some time $\hat{t}$, i.e. ${\boldsymbol\mu}^{B}_{i,t}(\hat{\theta})={\boldsymbol\mu_{i,t}}(\hat{\theta})$ for any $\hat{\theta} \in \Theta$ and $t \geq \hat{t}$. We then have
\begin{align}
{\boldsymbol\mu_{i,t}}(\hat{\theta}) \longrightarrow 0, \ \ \  \  \forall \hat{\theta}\in \Theta \setminus \mathcal{O}_i(\theta),
\end{align}
almost surely.
\end{mylemma}
Lemma \ref{Main Lemma} simply implies that the switching condition $\|{\boldsymbol\mu}^B_{i,t}(\cdot)-{\boldsymbol\mu}_{i,t-1}(\cdot) \|_{TV} < \tau$ is satisfied for all agents following a finite (but random) number of iterations. We also state the following proposition (using our notation) from \cite{moreau2005stability} to invoke later in the analysis. 
\begin{myproposition}\label{AUX}
Consider a sequence of directed graphs $\mathcal{G}_t = ([n],\mathcal{E}_t,A_t)$ for $t\in \mathbb{N}$ where $A_t$ is a stochastic matrix. Assume the existence of real numbers $\delta_{\max} \geq \delta_{\min}>0$ such that $\delta_{\min} \leq [A_t]_{ij}\leq \delta_{\max}$ for any $(i,j)\in \mathcal{E}_t$. Assume in addition that the graph $\mathcal{G}_t$ is bidirectional for any $t\in \mathbb{N}$. If for all $t_0\in \mathbb{N}$ there is a node connected to all other nodes across $[t_0,\infty)$, then the left product $A_tA_{t-1}\cdots A_1$ converges to a limit. 
\end{myproposition} We use the previous technical results to prove that under the proposed switching algorithm, all agents learn the truth, asymptotically and almost surely. 
\begin{mytheorem}[Learning in Switching Regimes]\label{mainTheorem}
Let the bound on log-marginals (assumption {\bf A1}), global identifiability of the true state (assumption {\bf A2}), and strong connectivity of the network (assumption {\bf A3}) hold. Then, following the updates in \eqref{eq:nonBayesianUpdate} and \eqref{nonbayes} using the switching rule in \eqref{SWITCHRULE}, all agents learn the truth exponentially fast with an asymptotic rate given by $\min_{\hat{\theta}\neq \theta}\{-\mathcal{I}(\hat{\theta},\theta)\}>0$.
\end{mytheorem}

Theorem \ref{mainTheorem} captures the trade-off between communication and informativeness of private signals. More specifically, private signals do not provide each agent with adequate information to learn the true state. Hence, agents require other signals dispersed throughout the network, which highlights the importance of communication. On the other hand, all-time communication is unnecessary since agents might only need a handful of interactions to augment their imperfect observations with those of their neighbors. 

 \section{Numerical Experiments}\label{sec:sim}

\begin{figure}[t]
\centering
\includegraphics[trim = 21mm 5mm 7mm 8mm, clip, scale=0.51]{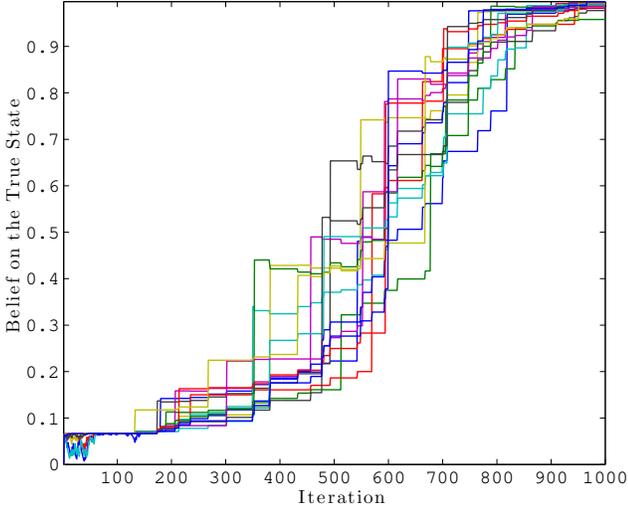}
\caption{The evolution of the belief on the true state for all agents in the network. Agents avoid all-time information exchange using the proposed switching rule, and eventually learn the truth.}
\label{Shekl}
\end{figure}

In this section, we exemplify the efficiency of the method using synthetic data. We generate a network of $n=15$ agents that aim to recover the true state $\theta=\theta_1$ among $m=16$ possible states of the world. The signals are binary digits, i.e., $\mathbf{s}_{i,t} \in \{0,1\}$ at each time $t$. For each agent $i\in [n]$, only state $i+1$ is {\it not} observationally equivalent to the true state $\theta_1$. This implies that $\Theta \setminus \mathcal{O}_i(\theta_1)=\{\theta_{i+1}\}$ which results in $\cap_{i=1}^n\mathcal{O}_i(\theta_1)=\{\theta_1\}$ to guarantee global identifiability of the true state. We set the threshold such that $\log_{10}\tau=-17$ for our switching rule and perform the updates \eqref{eq:nonBayesianUpdate} and \eqref{nonbayes} for $1000$ iterations. In Fig. \ref{Shekl}, we see that all agents reach consensus on the true state almost surely.

\begin{figure}[t]
\centering
\includegraphics[trim = 21mm 5mm 7mm 8mm, clip, scale=0.51]{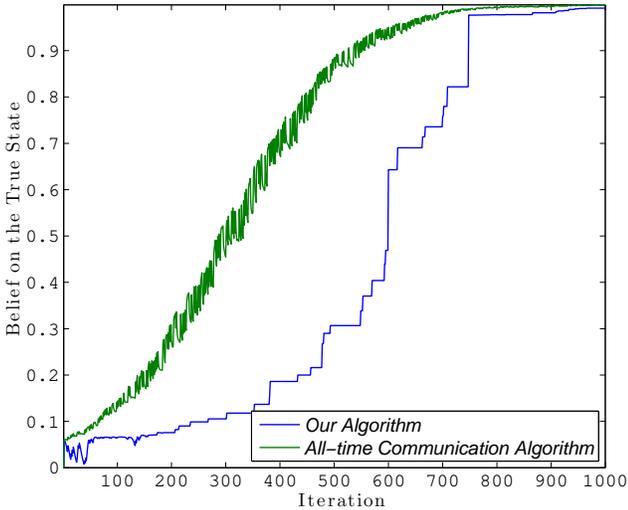}
\caption{The comparison of belief evolution for a randomly selected agent in the network. The blue curve is generated under the algorithm presented in this work, while the green one is based on the scheme in \cite{shahrampour2014distributed}.}
\label{shekl2}
\end{figure}

We now turn to compare the efficiency of the algorithm versus its counterpart in \cite{shahrampour2014distributed}. Fig. \ref{shekl2} represents the belief evolution under both algorithms for a randomly selected agent in the network. We observe that both algorithms converge; however, our proposed algorithm outperforms the one in \cite{shahrampour2014distributed} in terms of efficiency. The selected agent involves in interactions only $41$ times in $1000$ rounds. Therefore, the communication load simply reduces to $4.1\%$ comparing to the green curve, which proves a significant improvement.

\section{Concluding Remarks}\label{sec:conc}

In this paper we analyzed the problem of learning for a group of agents who try to infer an unknown state of the world. Agents rely on their private signals to perform a Bayesian update. However, private observations of a single agent may not provide sufficient information to identify the truth. Any time that private signals of agents lack adequate information, they engage in bidirectional communications with each other to benefit from side observations. We showed that under the proposed algorithm agents learn the true state asymptotically almost surely while dramatically saving on their communication budgets. Our future work focuses on the advancement of the proposed formulation by deriving optimality results in terms of the communication cost and convergence speed. This would in turn allow us to design an optimal informativeness threshold in the proposed switching strategies.


\section*{Appendix : Proofs}\label{omitted}

\textbf{\emph{Proof of Lemma \ref{Main Lemma}}}. Given the hypothesis, agent $i$ follows the Bayesian update after $\hat{t}$, and we have
\begin{align}
{\boldsymbol\mu}_{i,t}(\hat{\theta})={\boldsymbol\mu}^{B}_{i,t}(\hat{\theta}) = \frac{ {\boldsymbol\mu}_{i,t-1}(\hat{\theta})\ell_i(\mathbf{s}_{i,t}|\hat{\theta})} {\sum_{\tilde{\theta} \in \Theta}{\boldsymbol\mu}_{i,t-1}(\tilde{\theta})\ell_i(\mathbf{s}_{i,t}|\tilde{\theta})},
\end{align}
for any $\hat{\theta} \in \Theta$ and $t \geq \hat{t}$. Recalling that $\theta$ denotes the true state, we can write for any $t>\hat{t}$,
\begin{align}
\log \frac{{\boldsymbol\mu}_{i,t}(\hat{\theta})}{{\boldsymbol\mu}_{i,t}(\theta)}&=\log \frac{{\boldsymbol\mu}_{i,t-1}(\hat{\theta})}{{\boldsymbol\mu}_{i,t-1}(\theta)}+\log \frac{\ell_i(\mathbf{s}_{i,t}|\hat{\theta})}{\ell_i(\mathbf{s}_{i,t}|\theta)}. \label{eq:3333}
\end{align}
Therefore, for any $\hat{\theta} \in \mathcal{O}_i(\theta)$, we have
\begin{align}
\frac{{\boldsymbol\mu}_{i,t}(\hat{\theta})}{{\boldsymbol\mu}_{i,t}(\theta)}=\frac{{\boldsymbol\mu}_{i,\hat{t}}(\hat{\theta})}{{\boldsymbol\mu}_{i,\hat{t}}(\theta)},
\end{align}
for any $t>\hat{t}$ since in \eqref{eq:3333} the likelihood ratio is one, and $\log \frac{\ell_i(\mathbf{s}_{i,t}|\hat{\theta})}{\ell_i(\mathbf{s}_{i,t}|\theta)} = 0$ by definition of observationally equivalent states in \ref{def:observationallyEquivalnet}. On the other hand, for any $\hat{\theta} \in \Theta \setminus \mathcal{O}_i(\theta)$ simplifying \eqref{eq:3333} and dividing by $t$, we obtain
\begin{align}
\frac{1}{t}\log \frac{{\boldsymbol\mu}_{i,t}(\hat{\theta})}{{\boldsymbol\mu}_{i,t}(\theta)}&=\frac{1}{t}\log \frac{{\boldsymbol\mu}_{i,\hat{t}}(\hat{\theta})}{{\boldsymbol\mu}_{i,\hat{t}}(\theta)}+\frac{1}{t}\sum_{\tau=\hat{t}+1}^t\log \frac{\ell_i(\mathbf{s}_{i,\tau}|\hat{\theta})}{\ell_i(\mathbf{s}_{i,\tau}|\theta)}\\
&\longrightarrow \mathbb{E}\left[\log \frac{\ell_i(\cdot\vert \hat{\theta})}{\ell_i(\cdot \vert \theta)} \right]\\
&=-D_{KL}\left(\ell_i( \mathord{\cdot} |\theta) \| \ell_i( \mathord{\cdot} |\hat{\theta}) \right)<0,
\end{align} almost surely by the Strong Law of Large Numbers (SLLN). Note that since the signals are i.i.d. over time and the log-marginals are bounded (assumption {\bf A1}), SLLN could be applied. The above entails that ${\boldsymbol\mu}_{i,t}(\hat{\theta})\longrightarrow 0$ for any $\hat{\theta} \in \Theta \setminus \mathcal{O}_i(\theta)$, and thereby completing the proof.
 $\hfill \blacksquare $\\

\textbf{\emph{Proof of Theorem \ref{mainTheorem}}}. Fix any time $t_0\in \mathbb{N}$. When an agent uses Bayes' rule for $t\geq t_0$, in view of Lemma \ref{Main Lemma}, the condition $\|{\boldsymbol\mu}^{B}_{i,t}(\cdot)-{\boldsymbol\mu}_{i,t-1}(\cdot) \|_{TV} < \tau$ will be satisfied in a finite (random) time due to almost sure convergence of Bayes' rule. Therefore, all neighboring agents will eventually communicate with each other in the interval $[t_0,\infty)$. On the other hand, the underlying graph $\mathcal{G}$ is strongly connected by assumption {\bf A3}; hence, all the conditions of Proposition \ref{AUX} are satisfied and the left product $\mathbf{Q}_{t}\mathbf{Q}_{t-1}\cdots \mathbf{Q}_{1}$ has a limit, and since the matrices in the sequence $\{\mathbf{Q}_{t}\}_{t=1}^\infty$ are doubly stochastic by the switching rule in \eqref{SWITCHRULE}, we get
\begin{align}
\prod_{\rho=0}^{t-1} \mathbf{Q}_{t-\rho} \longrightarrow \frac{1}{n}\mathds{1}\mathds{1}^T, \label{convmatrix}
\end{align}
almost surely. Recalling \eqref{therightphi}, we can write 
\begin{align}
\frac{1}{t}\boldsymbol\phi_{i,t}(\hat{\theta}) &= \frac{1}{t}\sum_{\tau=0}^{t}\sum_{j=1}^n\left[\prod_{\rho=0}^{t-1-\tau}\mathbf{Q}_{t-\rho}\right]_{ij}\log \ell_j(\mathbf{s}_{j,\tau} |\hat{\theta})\\
&= \frac{1}{nt}\sum_{\tau=0}^{t}\sum_{j=1}^n\log \ell_j(\mathbf{s}_{j,\tau} |\hat{\theta})+\mathbf{e}_{i,t},\label{EQEQEQ}
\end{align}
where 
\begin{align}
\mathbf{e}_{i,t}=\frac{1}{t}\sum_{\tau=0}^{t}\sum_{j=1}^n\left(\left[\prod_{\rho=0}^{t-1-\tau}\mathbf{Q}_{t-\rho}\right]_{ij}-\frac{1}{n}\right)\log \ell_j(\mathbf{s}_{j,\tau} |\hat{\theta}).
\end{align}
Since the log-marginals are bounded (assumption {\bf A1}), in view of \eqref{convmatrix} we get 
\begin{align}
|\mathbf{e}_{i,t}|&\leq\frac{B}{t}\sum_{\tau=0}^{t}\sum_{j=1}^n\left|\left[\prod_{\rho=0}^{t-1-\tau}\mathbf{Q}_{t-\rho}\right]_{ij}-\frac{1}{n}\right|\\
&\longrightarrow 0, \label{EQEQEQEQ}
\end{align}
as $t\rightarrow \infty$, since Ces$\grave{\text{a}}$ro mean preserves the limit. Also, applying SLLN we have
\begin{align}
\frac{1}{nt}\sum_{\tau=0}^{t}\sum_{j=1}^n\log \ell_j(\mathbf{s}_{j,\tau} |\hat{\theta}) \longrightarrow \frac{1}{n}\sum_{j=1}^n\mathbb{E}\left[\log \ell_j(\cdot |\hat{\theta}) \right], 
\end{align}
almost surely. Combining above with \eqref{EQEQEQ} and \eqref{EQEQEQEQ} and recalling the definition of $\mathcal{I}(\hat{\theta},\theta)$ in \eqref{GAMMAdefinition}, we derive
\begin{align}
\frac{1}{t}\boldsymbol\phi_{i,t}(\hat{\theta})-\frac{1}{t}\boldsymbol\phi_{i,t}(\theta) \longrightarrow  \mathcal{I}(\hat{\theta},\theta),\label{rate}
\end{align}
almost surely, which guarantees that
\begin{align}
e^{\boldsymbol\phi_{i,t}(\hat{\theta})-\boldsymbol\phi_{i,t}(\theta)} \longrightarrow 0,  \label{EQ33}
\end{align}
for any $\hat{\theta}\in \Theta \setminus \{\theta\}$, since $\mathcal{I}(\hat{\theta},\theta)<0$ due to global identifiability of $\theta$ (assumption {\bf A2}). Now observe that 
\begin{align}
\boldsymbol\mu_{i,t}(\theta) &= \frac{\boldsymbol\mu_{i,0}(\theta)e^{\boldsymbol\phi_{i,t}(\theta)}}{\sum_{\tilde{\theta}\in\Theta}\boldsymbol\mu_{i,0}(\tilde{\theta})e^{\boldsymbol\phi_{i,t}(\tilde{\theta})}}\\
&=\frac{1}{1+\sum_{\tilde{\theta}\in \Theta \setminus \{\theta\}}\boldsymbol\mu_{i,0}(\tilde{\theta})e^{\boldsymbol\phi_{i,t}(\tilde{\theta})-\boldsymbol\phi_{i,t}(\theta)}}. \label{endLIMIT}
\end{align}
Taking the limit and using \eqref{EQ33}, the proof of convergence follows immediately, and per \eqref{rate} this convergence is exponentially fast with the asymptotic rate $\min_{\hat{\theta}\neq\theta }\{-\mathcal{I}(\hat{\theta},\theta)\}$ corresponding to the slowest vanishing summand in the denominator of  \eqref{endLIMIT}.  $\hfill \blacksquare $ \\
\bibliographystyle{IEEEtran}
\bibliography{BayesRef,shahin}

\begin{thebibliography}{10}
\providecommand{\url}[1]{#1}
\csname url@samestyle\endcsname
\providecommand{\newblock}{\relax}
\providecommand{\bibinfo}[2]{#2}
\providecommand{\BIBentrySTDinterwordspacing}{\spaceskip=0pt\relax}
\providecommand{\BIBentryALTinterwordstretchfactor}{4}
\providecommand{\BIBentryALTinterwordspacing}{\spaceskip=\fontdimen2\font plus
\BIBentryALTinterwordstretchfactor\fontdimen3\font minus
  \fontdimen4\font\relax}
\providecommand{\BIBforeignlanguage}[2]{{%
\expandafter\ifx\csname l@#1\endcsname\relax
\typeout{** WARNING: IEEEtran.bst: No hyphenation pattern has been}%
\typeout{** loaded for the language `#1'. Using the pattern for}%
\typeout{** the default language instead.}%
\else
\language=\csname l@#1\endcsname
\fi
#2}}
\providecommand{\BIBdecl}{\relax}
\BIBdecl

\bibitem{BorVar82}
V.~Borkar and P.~Varaiya, ``Asymptotic agreement in distributed estimation,''
  \emph{IEEE Transactions on Automatic Control}, vol. 27, no. 3, pp. 650--655,
  1982.

\bibitem{Tsit88}
J.~N. Tsitsiklis, ``Decentralized detection by a large number of sensors,''
  \emph{Mathematics of Control, Signals, and Systems}, vol. 1, no. 2, pp.
  167--182, 1988.

\bibitem{tsitsiklis1993decentralized}
J.~N. Tsitsiklis \emph{et~al.}, ``Decentralized detection,'' \emph{Advances in
  Statistical Signal Processing}, vol.~2, pp. 297--344, 1993.

\bibitem{jadbabaie2003coordination}
A.~Jadbabaie, J.~Lin, and A.~S. Morse, ``Coordination of groups of mobile
  autonomous agents using nearest neighbor rules,'' \emph{IEEE Transactions on
  Automatic Control}, vol.~48, no.~6, pp. 988--1001, 2003.

\bibitem{Olfati05}
R.~Olfati-Saber and J.~Shamma, ``Consensus filters for sensor networks and
  distributed sensor fusion,'' in \emph{IEEE Conference on Decision and
  Control}, 2005, pp. 6698 -- 6703.

\bibitem{tsitsiklis1984problems}
J.~Tsitsiklis, ``Problems in decentralized decision making and computation.''
  DTIC Document, Tech. Rep., 1984.

\bibitem{kar2012distributed}
S.~Kar, J.~Moura, and K.~Ramanan, ``Distributed parameter estimation in sensor
  networks: Nonlinear observation models and imperfect communication,''
  \emph{IEEE Transactions on Information Theory}, vol. 58, no. 6, pp.
  3575--3605, 2012.

\bibitem{atanasov2014distributed}
N.~A. Atanasov, J.~Le~Ny, and G.~J. Pappas, ``Distributed algorithms for
  stochastic source seeking with mobile robot networks,'' \emph{Journal of
  Dynamic Systems, Measurement, and Control}, 2014.

\bibitem{7040469}
N.~Atanasov, R.~Tron, V.~M. Preciado, and G.~J. Pappas, ``Joint estimation and
  localization in sensor networks,'' in \emph{IEEE Conference on Decision and
  Control (CDC)}, 2014, pp. 6875--6882.

\bibitem{jadbabaie2012non}
A.~Jadbabaie, P.~Molavi, A.~Sandroni, and A.~Tahbaz-Salehi, ``Non-bayesian
  social learning,'' \emph{Games and Economic Behavior}, vol.~76, no.~1, pp.
  210--225, 2012.

\bibitem{shahrampour2013online}
S.~Shahrampour, S.~Rakhlin, and A.~Jadbabaie, ``Online learning of dynamic
  parameters in social networks,'' in \emph{Advances in Neural Information
  Processing Systems}, 2013.

\bibitem{MosselSlyTamuz14}
E.~Mossel, A.~Sly, and O.~Tamuz, ``\BIBforeignlanguage{English}{Asymptotic
  learning on bayesian social networks},''
  \emph{\BIBforeignlanguage{English}{Probability Theory and Related Fields}},
  vol. 158, no. 1-2, pp. 127--157, 2014.

\bibitem{bullo2009distributed}
F.~Bullo, J.~Cort{\'e}s, and S.~Mart{\'\i}nez, \emph{Distributed control of
  robotic networks: a mathematical approach to motion coordination
  algorithms}.\hskip 1em plus 0.5em minus 0.4em\relax Princeton Univ Pr, 2009.

\bibitem{EgerMesBook}
M.~Mesbahi and M.~M.~Egerstedt, \emph{Graph theoretic methods in multiagent
  networks}.\hskip 1em plus 0.5em minus 0.4em\relax Princeton Univ Press, 2010.

\bibitem{Chamley2004}
C.~P. Chamley, \emph{{Rational Herds: Economic Models of Social
  Learning}}.\hskip 1em plus 0.5em minus 0.4em\relax Cambridge University
  Press, 2004.

\bibitem{Jackson2008}
M.~O. Jackson, \emph{Social and Economic Networks}.\hskip 1em plus 0.5em minus
  0.4em\relax Princeton, NJ, USA: Princeton University Press, 2008.

\bibitem{rad2010distributed}
K.~Rad and A.~Tahbaz-Salehi, ``Distributed parameter estimation in networks,''
  in \emph{IEEE Conference on Decision and Control (CDC)}, 2010, pp.
  5050--5055.

\bibitem{shahrampour2014distributed}
S.~Shahrampour, A.~Rakhlin, and A.~Jadbabaie, ``Distributed detection:
  Finite-time analysis and impact of network topology,'' \emph{arXiv preprint
  arXiv:1409.8606}, 2014.

\bibitem{lalitha2014social}
A.~Lalitha, A.~Sarwate, and T.~Javidi, ``Social learning and distributed
  hypothesis testing,'' in \emph{IEEE International Symposium on Information
  Theory (ISIT)}, 2014, pp. 551--555.

\bibitem{rahimian2014non}
M.~A. Rahimian, P.~Molavi, and A.~Jadbabaie, ``({N}on-) bayesian learning
  without recall,'' in \emph{IEEE Conference on Decision and Control (CDC)},
  2014, pp. 5730--5735.

\bibitem{shahin_CDC_2013}
S.~Shahrampour and A.~Jadbabaie, ``Exponentially fast parameter estimation in
  networks using distributed dual averaging,'' in \emph{IEEE Conference on
  Decision and Control (CDC)}, 2013, pp. 6196--6201.

\bibitem{nedic2014nonasymptotic}
A.~Nedi{\'c}, A.~Olshevsky, and C.~A. Uribe, ``Nonasymptotic convergence rates
  for cooperative learning over time-varying directed graphs,'' \emph{arXiv
  preprint arXiv:1410.1977}, 2014.

\bibitem{CoverThomas1991}
T.~M. Cover and J.~A. Thomas, \emph{Elements of Information Theory}.\hskip 1em
  plus 0.5em minus 0.4em\relax Wiley Series in Telecommunications, 1991.

\bibitem{moreau2005stability}
L.~Moreau, ``Stability of multiagent systems with time-dependent communication
  links,'' \emph{IEEE Transactions on Automatic Control}, vol.~50, no.~2, pp.
  169--182, 2005.

\end{thebibliography}

\end{document}